\documentclass[11pt,a4paper]{article}
\usepackage[hyperref]{acl2021}
\usepackage{times}
\usepackage{latexsym}

\usepackage{enumitem}
\usepackage{amsmath, amssymb}
\usepackage{pgfplots}
\usepackage{multicol}
\usepackage{booktabs}
\usepackage{soul}

\newcommand{\Var}{\text{Var}}
\newcommand{\Expect}{\mathbb{E}}

\usepackage{microtype}

\aclfinalcopy 
\def\aclpaperid{***} 

\title{The statistical advantage of automatic NLG metrics at the system level}

\author{
      Johnny Tian-Zheng Wei \hspace{5em} Robin Jia \\
      Department of Computer Science, University of Southern California \\
      \texttt{\{jtwei, robinjia\}@usc.edu}
}
\date{6/1/2021}

\begin{document}
\maketitle
\begin{abstract}
Estimating the expected output quality of generation systems is central to NLG. This paper qualifies the notion that automatic metrics are not as good as humans in estimating system-level quality. Statistically, humans are unbiased, high variance estimators, while metrics are biased, low variance estimators. We compare these estimators by their error in pairwise prediction (which generation system is better?) using the bootstrap. Measuring this error is complicated: predictions are evaluated against noisy, human predicted labels instead of the ground truth, and metric predictions fluctuate based on the test sets they were calculated on. By applying a bias-variance-noise decomposition, we adjust this error to a noise-free, infinite test set setting. Our analysis compares the adjusted error of metrics to humans and a derived, perfect segment-level annotator, both of which are unbiased estimators dependent on the number of judgments collected. In MT, we identify two settings where metrics outperform humans due to a statistical advantage in variance: when the number of human judgments used is small, and when the quality difference between compared systems is small.\footnote{The data and code to reproduce our analyses are available at \url{https://github.com/johntzwei/metric-statistical-advantage}.}
\end{abstract}

\pgfmathdeclarefunction{gauss}{2}{%
  \pgfmathparse{1/(#2*sqrt(2*pi))*exp(-((x-#1)^2)/(2*#2^2))}%
}

\begin{figure}[t]
    \centering
\begin{tikzpicture}
\begin{axis}[
  no markers, domain=-2:5, samples=100,
  axis lines*=middle, xlabel=\empty, ylabel=$0$,
  every axis y label/.style={at=(current axis.below origin),anchor=north},
  every axis x label/.style={at=(current axis.right of origin),anchor=west},
  height=5cm, width=9cm,
  xtick=\empty, ytick=\empty,
  enlargelimits=false, clip=false, axis on top,
  grid = major,
  ] 
  
  \addlegendimage{empty legend}
  
  \addplot [very thick,orange!50!white] {gauss(1.5,0.7)};
  \addplot [very thick,cyan!50!black] {gauss(2.7,2)};
  
  \addplot [fill=cyan!20, draw=none, domain=-2:0] {gauss(2.7,2)} \closedcycle;
  \addplot [fill=orange!20, draw=none, domain=-2:0] {gauss(1.5,0.7)} \closedcycle;
  
  \addplot [dashed] coordinates {(1.5,0) (1.5, 0.57)};
  \addplot [dashed] coordinates {(2.7,0) (2.7, 0.2)};
  
  \addplot [very thick,orange!50!white] {gauss(1.5,0.7)};
  \addplot [very thick,cyan!50!black] {gauss(2.7,2)};
  
  \draw [yshift=.3cm, latex-latex](axis cs:1.5,0) -- node [fill=white] {\small Bias} (axis cs:2.7,0);
  \draw [yshift=-0.4cm, latex-latex](axis cs:2.7,0) node [fill=white] {$\delta^H$};
  \draw [yshift=-0.4cm, latex-latex](axis cs:4.15,0) node [fill=white] {\small (True difference)};
 
  \legend{}
  
  \addlegendentry{\small \hspace{-.6cm} Dist. of estimator}
  \addlegendentry[align=left]{\small Metric ($\widehat{\delta^M}$)}
  \addlegendentry[align=left]{\small Human ($\widehat{\delta^H}$)  }
\end{axis}

\end{tikzpicture}
    \caption{Distribution of estimators for the true difference in system quality $\delta^H$ between two generation systems (for illustrative purposes). Notation is defined in \S \ref{section:formal_pairwise_judgments}. An estimate incurs prediction error if its sign is opposite to the true difference. While humans provide an unbiased estimator of the difference, a biased estimator derived from a metric can have a smaller error probability (shaded areas) due to its lower variance. Evidence supporting the illustration can be found in \S \ref{section:comparing_to_humans}.} \label{figure:pairwise_illustration}
\end{figure}
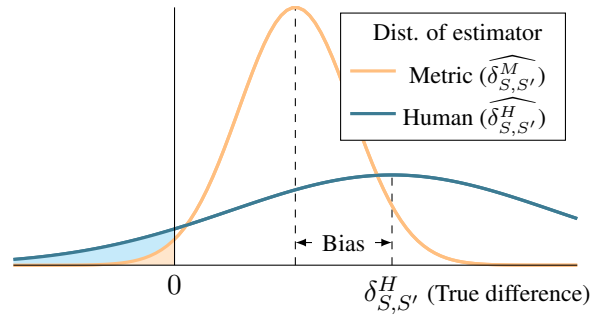

\setlength{\abovedisplayskip}{5pt}
\setlength{\belowdisplayskip}{5pt}

\setitemize{topsep=0pt,itemsep=1ex,partopsep=0ex,parsep=0ex, left=0pt..0.7em}


\section{Introduction}

Automatic metrics are involved in many developmental settings for natural language generation (NLG) systems. In machine translation (MT), metrics like BLEU \cite{papineni-etal-2002-bleu} enable settings where the amount of human effort required would be infeasible, such as architecture or hyperparameter search \cite{britz-etal-2017-massive}.  As objective, reproducible quantities, BLEU scores facilitate cross-paper comparisons \cite{post-2018-call}. Historically, progress in MT has been attributed to its use \cite{callison-burch-etal-2006-evaluating}. Metrics are an active research area in many NLG subfields, including summarization \cite{lin-2004-rouge}, dialogue \cite{DBLP:conf/aaai/TaoMZY18}, and image captioning \cite{DBLP:conf/eccv/AndersonFJG16}, which seek to realize the goal of quick and reliable automatic evaluation.

In all these subfields, the primary goal when conducting evaluation is typically to compare NLG systems.
Both human annotators and automatic metrics produce \emph{segment-level} scores, i.e., scores for individual examples,
so comparing systems requires aggregating segment-level scores into an overall \emph{system-level} score for each system. Ideally, we would compare systems by their expected human annotator score (an average over infinite human judgments), which we term the true system quality. In practice, we can only estimate this expectation with a sample mean over a finite number of human judgments. Metrics offer a cheaper alternative: we can instead compare systems by their aggregate metric scores on a number of system outputs. When comparing systems, we care primarily about how well we estimate the \emph{difference} of their true system qualities, and in particular the sign of this difference (i.e., which system is better), which we term the true pairwise label.


There is a gap in our understanding of \emph{system-level} metrics. To recount a perplexing anecdote, in the most recent edition of the WMT metrics shared task \cite{mathur-EtAl:2020:WMT}, initial human evaluation disagreed with most metrics on a pairwise prediction of two translation systems. In a manual re-evaluation, the second round results favored the metrics. Our paper offers a statistical explanation for how humans could go ``wrong'': even if human estimation for the difference in system quality is unbiased, it has high variance. On the other hand, while estimators based on metrics are biased, they have low variance. It is therefore possible for metrics to give a more accurate pairwise prediction than humans when the bias is small (see illustration in Figure \ref{figure:pairwise_illustration}). Our paper explores this distinction in the following three questions:

\textbf{(1)} \emph{How can we evaluate system-level metrics?} When observing estimator error in terms of pairwise predictions, predictions are evaluated against noisy, human predicted labels rather than the ground truth. In addition, metric predictions fluctuate based on the sample of outputs from the generation system. To disentangle these properties, we examine observed estimator error under a bias-variance-noise decomposition. \textbf{Under simulation, we find that the label noise and metric variance account for a small fraction of observed error in both MT and summarization.}

\textbf{(2)} \emph{How good are these metrics?} We compare the errors of metric estimators computed on an infinite number of system outputs, against human estimators with varying amounts of human judgment. We also derive the error of a perfect segment-level annotator (i.e. they provide noiseless/expected human scores for each output), which is also unbiased and judgment dependent. \textbf{Empirically, some MT metrics exceed the performance of unbiased estimators with a small number of judgments.}

\textbf{(3)} \emph{What are the limits of system-level evaluation?} The perfect segment-level annotator, as the noiseless human, provides an optimistic estimate for the number of human judgments necessary to achieve a fixed performance. With a power analysis, we can analytically calculate the number of judgments necessary to detect differences between systems of varying sizes. \textbf{When differences in system quality are small, a prohibitively large number of perfect annotator judgments are required to give a correct pairwise prediction.}

\section{Formalization} \label{section:formalization}

\subsection{System-level scores} \label{section:formalization_system_scores}

We will now formalize scoring at the system level, adopting notation from \citet{chaganty-etal-2018-price}. Let $\mathcal{X}$ be a distribution over inputs (e.g. source sentences), and $\mathcal{S}$ be a set of systems (e.g. all translation systems in WMT). Each system $S\in\mathcal{S}$ takes input $x\sim\mathcal{X}$ and returns output $z = S(x)$ (e.g. $z$ is a translation). Let $H(z)$ be a random variable representing a human judgment according to some evaluation prompt (e.g. translation adequacy, from 0-100). A central quantity of interest is the quality of system $S$, defined as
\begin{equation}\label{eq:true_quality}
    \mu^{H}_{S} = \Expect_{x\sim\mathcal{X}}[H(S(x))]
    \end{equation}
and is not directly observable as it requires infinite human judgment. We can estimate (\ref{eq:true_quality}) with a finite test set of $n$ examples. Let $x^{(1)}, \dots, x^{(n)} \overset{\text{i.i.d.}}{\sim} \mathcal{X}$ be a sampled test set and $z^{(1)}, \dots, z^{(n)}$ be the set of outputs where each $z^{(i)} = S(x^{(i)})$. Human judgments are sampled independently as $y^{(i)} \sim H(z^{(i)})$. The sample mean
\begin{equation} \label{eq:sample_quality}
    \widehat{\mu^{H}_{S}} = \frac{1}{n}\sum_{i=1}^{n} y^{(i)}
\end{equation}
is an unbiased estimator of \eqref{eq:true_quality}. Only \eqref{eq:sample_quality} is observable, which is a noisy approximation of \eqref{eq:true_quality}. 

A cheaper alternative to estimating the true quality scores is with an estimator based on an automatic metric. Let $M$ (e.g. {\sc BERTscore}) be an automatic metric that takes as input any number of outputs from a system $S$ and produces score
\begin{equation}
    \widehat{\mu^M_S} = M(z^{(1)}, \dots, z^{(n)})
\end{equation}
where $\widehat{\mu_{S}^M}$ is a biased estimator of $\mu^{H}_{S}$. As the test set is sampled, the metric score has non-zero variance. Note that while we use the greek letter $\mu$, only some system-level metrics (e.g. {\sc ROUGE}) are averages of their segment-level counterparts (their score decomposes to $\widehat{\mu^M_S} = \frac{1}{n}\sum_{i=1}^n M(z^{(i)})$). Empirically, we find that metrics using other aggregation strategies have convergent properties similar to an average (see Appendix \ref{section:metric_convergence}). We sidestep this by defining the ``true'' metric score as
\begin{equation}
    \mu^M_S = M(z^{(1)}, \dots, z^{(m)})
\end{equation}
for test sets of size $m$ sufficiently large so that this true score is nearly constant.

\subsection{Problems in evaluating with correlation}

Research in system-level metrics have a tradition of evaluating metric correlation to human judgment with the Pearson correlation coefficient \cite{reiter-2018-structured}. Formally, these evaluations compare
$\widehat{r_M} = \text{Corr}_{\mathcal{S}}(\widehat{\mu^H_S}, \widehat{\mu^M_S})$
for different metrics $M$. 

Recently, \citet{mathur-etal-2020-tangled} highlights two issues with the use of correlation: 
First, Pearson's $r$ is neither interpretable nor reflective of system-level metric use in practice. 
Second, outlier systems (systems with very high/low human/metric scores) can arbitrarily inflate Pearson's $r$, and outlier systems often exist. 
\citet{mathur-etal-2020-tangled} propose evaluating metric accuracy in pairwise prediction (can the metric differentiate which generation system is better?) as an alternative that mitigates the issues mentioned above. 

We add two points that apply to any measure of metric performance, correlation or pairwise predictions: 
First, metrics cannot be perfect due to noise in human labels. For instance, while $r$ ranges from $[-1,1]$, even for the metric that predicts $\mu^H_S$ it has $\text{Corr}_{\mathcal{S}}(\widehat{\mu^H_S}, \mu^H_S) < 1$ due to noise in $\widehat{\mu^H_S}$. It is unclear what is the true upper bound of performance we can expect to achieve.
Second, direct measurement of any performance measure on our datasets introduces sample bias \cite{DBLP:journals/corr/abs-2005-09619}. For correlation, $\widehat{r_M}$ could be high because $\widehat{\mu^H_S}$ and $\widehat{\mu^M_S}$ happened to align for this data collection, but a repeat experiment could yield different results. A more holistic view is to give an estimate of average case performance.\footnote{Pearson's $r$ was not formulated for individual distributions $\widehat{\mu^H_S}$ and $\widehat{\mu^M_S}$ for each datapoint, so applying the William's test \cite{graham-baldwin-2014-testing} also falls short here.}

The evaluation methodology we derive in \S \ref{section:decomposing_error} addresses the latter points we raise for pairwise predictions and mean squared error (which has direct relationship to the correlation). However, we also believe that pairwise predictions is a step in the right direction, and our discussion continues with pairwise predictions.

\subsection{Pairwise predictions} \label{section:formal_pairwise_judgments}
We will now formalize pairwise predictions. For systems $S, S' \in \mathcal{S}$, define the true difference in their system scores as 
\begin{equation} \label{eq:true_diff_label}
    \delta^H = \mu^H_{S} - \mu^H_{S'}  
\end{equation}
and the observed difference as
\begin{equation}
    \widehat{\delta^H} = \widehat{\mu^H_{S}} - \widehat{\mu^H_{S'}} 
\end{equation}
and likewise for the differences $\delta^M$ and $\widehat{\delta^M}$ w.r.t. to a metric $M$. In practice, we are interested in the pairwise prediction of $S$ and $S'$ i.e. whether $\delta^H \stackrel{?}{>} 0$, given that we have collected human judgments (we observe $\widehat{\delta^H} \lessgtr 0$), or computed metric scores (we observe $\widehat{\delta^M} \lessgtr 0$). Refer to Figure \ref{figure:pairwise_illustration} for an illustration.

To operationalize the pairwise prediction of $S$ and $S'$, let the true pairwise label
\begin{equation}
    \Delta^H = \text{sign}(\delta^H)
\end{equation}
be defined as the central quantity of interest. Define the human predicted pairwise label as
\begin{equation}
    \widehat{\Delta^H} = \text{sign}(\widehat{\delta^H})
\end{equation}
and likewise for the true and estimated predictions $\Delta^M$ and $\widehat{\Delta^M}$ w.r.t. to a metric M. The 0-1 classification loss for metric $M$ on this example is
\begin{equation}
    L(\Delta^H, \widehat{\Delta^M}) = \mathbb{I}[\Delta^H \neq \widehat{\Delta^M}]
\end{equation}
and the pairwise error of an estimator is the loss incurred averaged over all pairwise examples. Ideally, we could calculate the true error of $M$
\begin{equation}  \label{eq:true_pairwise_err}
    \text{Err}_{\text{true}}(M) = \mathbb{E}_{\mathcal{S}}[L(\Delta^H, \Delta^M)]
\end{equation}
but we can only compute an error of $M$ with noisy human labels and metric scores estimated from finite sized test sets
\begin{equation} \label{eq:avg_pairwise_err}
    \text{Err}_{\text{obs}}(M) = \mathbb{E}_{\mathcal{X}, \mathcal{S}}[L(\widehat{\Delta^H}, \widehat{\Delta^M)}]
\end{equation}
which is typically estimated when we calculate metric pairwise accuracy from our datasets.

\begin{table*}[!t]
    \small
    \begin{minipage}{.5\textwidth}
    \begin{tabular}{r c | c c c }
    \toprule
    & & \multicolumn{3}{c}{Error components} \\
    & $\text{Err}_{\text{obs}}(\cdot)$ & $c_0\text{Noise}$ & $\text{Bias}$ & $c_1\text{Var}$  \\
    \midrule
Optimal ($\Delta^{H*}$) & 0.047 & 0.000 & 0.000 & \bf{0.047} \\
Human ($\widehat{\Delta^{H}}$) & 0.065 & 0.019 & 0.000 & \bf{0.047} \\
\sc{BERTscore} & 0.102 & 0.003 & \bf{0.086}& 0.013 \\
\sc{chrF} & 0.124 & 0.010 & \bf{0.105}& 0.009 \\
\sc{BLEURT}** & 0.128 & 0.005 & \bf{0.108}& 0.016 \\
\sc{BLEU} & 0.141 & 0.008 & \bf{0.127}& 0.007 \\
\sc{TER} & 0.184 & 0.002 & \bf{0.173}& 0.009 \\
\bottomrule
    \end{tabular}
    \end{minipage}
    \begin{minipage}{.5\textwidth}
    \begin{tabular}{r c | c c c }
    \toprule
    & & \multicolumn{3}{c}{Error components} \\
    & $\text{Err}_{\text{obs}}(\cdot)$ & $c_0\text{Noise}$ & $\text{Bias}$ & $c_1\text{Var}$  \\
    \midrule
Optimal ($\Delta^{H*}$) & 0.045 & 0.000 & 0.000 & \bf{0.045} \\
Human ($\widehat{\Delta^{H}}$) & 0.067 & 0.022 & 0.000 & \bf{0.046} \\
\sc{ROUGE} & 0.296 & -0.006 & \bf{0.294}& 0.008 \\
\sc{METEOR} & 0.296 & 0.004 & \bf{0.287}& 0.005 \\
\sc{ROUGE-WE} & 0.317 & 0.007 & \bf{0.301}& 0.008 \\
\sc{BERTscore} & 0.330 & -0.004 & \bf{0.338}& -0.004 \\
\sc{SUPERT}*** & 0.390 & 0.000 & \bf{0.382}& 0.008 \\
\bottomrule
    \end{tabular}%
    \end{minipage}
    \caption{Decomposition of the pairwise error of different metrics (left: WMT, right: SummEval). Highlighted in bold is the largest error component. 10K boostrap trials are conducted for estimation of the expectations (estimation error $<10^{-3}$). *Denotes an estimator assumed to be unbiased in the simulation. **{\sc BLEURT} is evaluated only on WMT2019. ***{\sc SUPERT} is a reference-less metric.}
    \label{table:bvd_wmt}
\end{table*}

\section{Datasets}


\subsection{WMT16-19 metrics shared task}

 \textbf{Data.} We use the past 4 years of to-English translation data from the WMT metrics shared task \cite{bojar-etal-2016-results, bojar-etal-2017-results, ma-etal-2018-results, ma-etal-2019-results}.\footnote{The WMT20 metrics shared task data was not publicly available at the time of submission.} Across all years and language pairs, there are 261 MT systems. Pairs of MT systems are extracted within each year, within each language pair, resulting in 1324 pairwise examples. For each output of an MT system, there are one or more humans judgements and one reference for metric scoring. 1306-5117 outputs were collected for each MT system totaling about 1312-5612 judgments, depending on the year and language pair. For ease of interpretation, we always use raw direct assessment judgments which range from 0-100. 
 

\textbf{Metrics.} We evaluate the performance of the three metrics included in SacreBleu \cite[BLEU, TER, chrF;][]{post-2018-call, koehn-etal-2007-moses}. These three have also participated in every year of the metrics task as baselines. In addition, we include two recently developed metrics: {\sc BERTscore} \cite{DBLP:conf/iclr/ZhangKWWA20} and BLEURT \cite{sellam-etal-2020-bleurt}. Both metrics are found to effectively utilize contextual embeddings \cite{devlin-etal-2019-bert}, and BLEURT is a learned metric (tuned on data outside of WMT2019). For all metrics, we use the default settings for scoring. Since BLEURT is trained on WMT15-18, we test it only on WMT2019 pairs.

\subsection{SummEval}

\textbf{Data.} The SummEval dataset \cite{fabbri2020summeval} contains 100 outputs from 17 summarization systems. This results in 136 pairwise examples. For each system output, 3 expert judgments, and 11 references for metric scoring. Each summarization is judged in four categories from 0-5: coherence, consistency, fluency, and relevance. To compute system-level human scores for a system, we first average over categories for an aggregate expert score, and then average the aggregated expert scores per system. Metric scores for system outputs were computed against as many references as possible.

\textbf{Metrics.} We evaluate the performance of several metrics that were found to be effective at the system-level in \citet{fabbri2020summeval}. This includes the traditional ROUGE-4 \cite{lin-2004-rouge} summarization metric, its extension ROUGE-WE \cite{ng-abrecht-2015-better}, and METEOR \cite{lavie-agarwal-2007-meteor}. In addition, we include two metrics based on BERT \cite{devlin-etal-2019-bert}. BertScore \cite{DBLP:conf/iclr/ZhangKWWA20}, also present in the WMT analysis, and SUPERT \cite{gao-etal-2020-supert}, which is a reference-less metric for summarization.

\section{Decomposing observed metric error} \label{section:decomposing_error}

Two sources of variation distinguish the observed pairwise error \eqref{eq:avg_pairwise_err} from the true error in \eqref{eq:true_pairwise_err} --- the noise in the human predicted labels due to finite judgements, and the variance in the metric due to finite test sets. Approximating \eqref{eq:avg_pairwise_err} is straightforward with the bootstrap, but disentangling the error from these two sources of variation requires more care. With the bias-variance-noise decomposition, we can adjust our observed error estimates to the noise-free, infinite test set setting of the true error.

\subsection{The bias-variance-noise decomposition}
The bias-variance-noise decomposition due to \citet{DBLP:conf/icml/Domingos00a} decomposes the observed pairwise error in \eqref{eq:avg_pairwise_err} w.r.t. two constant labels for any pairwise example on systems $S, S' \in \mathcal{S}$:
\begin{itemize}
    \item The \emph{true pairwise label} for this example is
    \begin{align} \label{eq:optimal_prediction}
        \Delta^{H*} := \arg\min_{y \in \{-1, 1\}} \Expect_{\mathcal{X}} [ L(\widehat{\Delta^H}, y)]
    \end{align}
    and the estimator that produces these true labels has, by definition, the lowest observed error. In the decomposition, the human predicted label noise and metric bias is defined relative to the true labels. Assuming the central limit theorem (proof in Appendix \ref{appendix:optimal_predictions_proof}), we actually have $\Delta^{H*} = \Delta^{H}$ as defined in eq. \eqref{eq:true_diff_label}.
    
    \item The \emph{main prediction} of a metric for this ex. is
    \begin{align}
        \Delta^{M*} = \arg\min_{y \in \{-1, 1\}} \Expect_{\mathcal{X}} [ L(\widehat{\Delta^M}, y)]
    \end{align}
    and we assume that the metric prediction converges onto the main prediction as the test data increases for $S$ and $S'$ (empirically validated in Appendix \ref{section:metric_convergence}). In the decomposition, the metric variance is defined relative to the main prediction.
\end{itemize}

\noindent Starting from the loss incurred by $M$ on this pairwise example, the decomposition gives us 
\begin{align}
    \Expect_{\mathcal{X}}[L(\widehat{\Delta^H}, \widehat{\Delta^M})] &= c_0 \text{Noise}(\widehat{\Delta^H}) \\
    &\hspace*{-1cm}+\text{Bias}(\widehat{\Delta^M}) + c_1 \text{Var}(\widehat{\Delta^M}) \nonumber
\end{align}
where
\begin{itemize}
    \item $\text{Noise}(\widehat{\Delta^H}) = \Expect[L(\widehat{\Delta^H}, \Delta^{H*})]$ where the noise is an irreducible loss incurred by computing pairwise accuracy to the human predicted labels instead of the true labels. Note that this noise term also exactly corresponds to the lowest achievable observable error (see \S \ref{section:upper_bound}).
    
    \item $\text{Bias}(\widehat{\Delta^M}) = L(\Delta^{H*}, \Delta^{M*})$ where the bias is 0 if the main prediction is correct (w.r.t. to the true label), and 1 otherwise. Note that this term is also the true error of a metric estimator in a noise-free, infinite test set setting. For unbiased estimators this term is zero, as their main prediction matches the true label.
    
    \item $\Var(\widehat{\Delta^M}) = \Expect[L(\widehat{\Delta^M}, \Delta^{M*})]$ where the variance is a likelihood that the estimator deviates from its main prediction under random sampling.
\end{itemize}
\begin{itemize}
    \item $c_0 = 2P_{\mathcal{X}}(\widehat{\Delta^M} = \Delta^{H*}) - 1$ which means that the influence of label noise on the error becomes small if the estimator prediction are close to random chance. When the estimator gives constant predictions, the sign of $c_0$ is dependent on the estimator's correctness.
    
    \item $c_1 = 1$ if $\widehat{\Delta^M} = \Delta^{H*}$ and $c_1 = -1$ otherwise. Variance can both increase and decrease the observed error. If the estimator is unbiased, the variance causes the prediction to from the correct main prediction. On the other hand, for a biased estimator, deviation from its incorrect main prediction occasionally decreases the error.
\end{itemize}
Unlike the decomposition for mean squared error, the interaction between the $c_0$ and $\text{Var}$ terms only allows the error of two hypothetical settings to be read off directly from the table: when $\text{Noise}\xrightarrow[]{} 0$, corresponding to estimator error when computed against the ground truth; or when $\text{Noise} + \text{Var} \xrightarrow[]{} 0$, when the ground truth is used and metrics have access to an infinite test set for scoring.

\subsection{A lower bound for the observed error} \label{section:upper_bound}

By definition the constant estimator that produces the true pairwise labels $\Delta^{H*}$ (defined in \eqref{eq:optimal_prediction}) for each pairwise example has the lowest possible observable error. The observable error of this optimal estimator is exactly $\Expect[L(\widehat{\Delta^H}, \Delta^{H*})] =  \text{Noise}(\widehat{\Delta^H})$. Since this estimator is constant it has no variance, and since it is instantiated by definition it has no bias. Analytically, the observed error of any estimator is lower bounded by $\text{Noise}(\widehat{\Delta^H})$ and is the agreement of our human predicted labels with the ground truth.

\subsection{Best-faith estimation with the bootstrap} \label{section:best_faith_estimation}

Assuming the bootstrap \cite{DBLP:books/sp/EfronT93} which is a common procedure in NLP \cite{dror-etal-2018-hitchhikers}, we can estimate the expectation quantities in the decomposition. By assuming that sampling with replacement from our datasets approximates real sampling, we can repeatedly simulate the quantity in an expectation. Taking the mean over trials gives the bootstrap estimate of the expectation. We emphasize that this is a regular application of a widely accepted technique---the bootstrap assumption allows us to study problems that would be impossible due to the cost of repeat experiments. 


\subsection{Results} \label{section:decomposition_analysis}

The following analyses refer to the error components (averaged over all examples) from the simulated decomposition presented in Table \ref{table:bvd_wmt}.


\textbf{The noise component almost always accounts for a small fraction of the total error.} We found this to be counterintuitive---while the lowest observable error (optimal predictions, see \S \ref{section:upper_bound}) incur about $5\%$ error on both datasets, the influence of the noise is much smaller than those errors suggest. For the constant $c_0$ scaling the noise, $c_0=0$ if the metric prediction is near random. Since the $c_0\text{Noise}$ term on average is small two cases hold true: when humans are uncertain about the example (noise term large) metrics are as well ($c_0$ term small), and when metrics are certain about the examples ($c_0$ term large) humans are as well (noise term small). The second case empirically shows studying the sampling distribution of metrics \cite{koehn-2004-statistical, berg-kirkpatrick-etal-2012-empirical} is effective, as metric certainty in the difference of system quality often implies human certainty.

\textbf{Metric variance introduces little to the pairwise error, because it is low.} Alternatively, metrics stand to gain little from using more test set examples. In MT, dropping both the noise and variance components for the error results in at most a 1 or 2 percent reduction in the observed error (see \S 9 for the implications in metrics research). Metrics generally have low variance, so at the test set sizes of WMT and SummEval, they are likely to converge to their main predictions.



\begin{figure}[t]
    \includegraphics{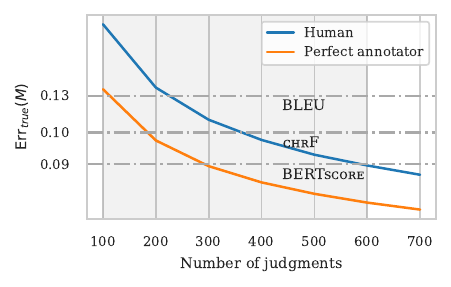}
\caption{Comparison of metrics to human and perfect annotator estimators with varying number of judgments in WMT. Errors are adjusted to an idealized setting where true predictions are used for evaluation and metrics are computed on infinite test sets; here metric predictions become constant, so their errors are constant. Shaded in grey is the region where {\sc BERTscore} is superhuman. Results for SummEval are in Appendix \ref{appendix:power_analysis}.}
\label{fig:human_comparisons}
\end{figure}


\section{Comparing to the human estimator} \label{section:comparing_to_humans}


In \S \ref{section:decomposing_error}, several MT metrics approach the error of the WMT human evaluation. The WMT human evaluation is expensive, using thousands of judgments per translation system. While each human judgment has associated monetary cost, once a large test set is collected, running metrics only incurs computational cost. This section explores this asymmetry, and seeks to understand how much metric predictions are worth, in terms of human judgments.

\subsection{Noise-free, variance-free error estimates}

We wish to give our best comparison between metrics and unbiased estimators (humans or the perfect annotator). Ideally, metrics would be given their best chance to perform, by using an infinite test set. With the decomposition, we can adjust metric errors estimates to a noise-free and infinite test set setting by taking only their bias component. For human and perfect annotator estimators, we can adjust their errors to a noise-free setting by taking only the variance component. The following sections compare these adjusted errors.

\subsection{Simulating the perfect annotator}

While we can estimate the lower bound to the pairwise error for a given dataset (in \S \ref{section:upper_bound}), it is achieved by a constant estimator using system-level ground truth. Comparing segment-level metrics against the unbiased ``perfect annotator'', or the best scorer at the segment-level, is more informative. At the high-level, we can simulate scoring with the perfect annotator at $n$ judgments using the human estimator at $n' > n$ judgments to match the variance of the perfect annotator estimator.

Let's start from the unbiased human estimator $\widehat{\mu_S^H}$ \eqref{eq:sample_quality}. Recall that the estimator is a sample mean, so its variance is $\Var(\widehat{\mu_{S}^H}) = \Var(H(x)) / n$.
An insight from \citet{chaganty-etal-2018-price} gives us the decomposition of the variance of $H(x)$
\begin{align} \label{law_of_tv}
    \Var(H(x)) &= \Var(\Expect [H(x) | x]) \\
    &+ \Expect[\Var(H(x) | x)] \nonumber
\end{align}
with the law of total variance. In words, the \emph{variance} term can be thought of as the variance of each output sentence's true quality score (some translations produced by $S$ are better than others) and the \emph{expectation} term is the noise introduced by the humans when estimating the quality of a sentence (human scores have mean 0 noise around an output's true quality score). 

One intuition is that even if a perfect annotator gives the correct score for each sentence, every time, there is still some unavoidable variance in the estimator due to the variance of the hypothetical quality scores for each output. To formalize this notion, let $P(x) = \mathbb{E}[H(x)|x]$ be the human scoring function of a ``perfect annotator'', and the estimator $\widehat{\mu_S^P}$ be an empirical mean of $n$ independent samples from $P(x)$ similar to eq. \eqref{eq:true_quality}. As a sample mean, $\Var(\widehat{\mu_S^P}) = \Var(P(x))/n$. Relating this to \eqref{law_of_tv}
\begin{equation}
    \Var(H(x)) = \Expect[\Var(H(x) | x)] + \Var(P(x))
\end{equation}
and while $\Var(P(x))$ is not directly observable, we can calculate $\Var(H(x))$ with the sample variance on all the human judgments, and $\Expect[\Var(H(x) | x)]$ with a pooled variance over variances from repeat human judgments on the same output sentence.

Our final step considers the efficiency ratio $r = \Var(H(x))/\Var(P(x))$. If we are interested in the perfect annotator estimator at $n$ judgments, the human estimator at $n'=rn$ judgments has variance
\begin{align}
    \Var(\widehat{\mu_{S}^H}) &= \frac{\Var(H(x))}{rn} \\ &=  \frac{\Var(P(x))}{n} = \Var(\widehat{\mu_{S}^P})
\end{align}
and we invoke the central limit theorem to claim both $\widehat{\mu_{S}^P}$ and $\widehat{\mu_{S}^H}$ are normal. This completes our reasoning that for scoring on the system-level, sampling $n' = nr$ human judgments is nearly equivalent to sampling $n$ perfect annotator judgments. See Appendix \ref{appendix:perfect_annotator_derivation} for step-by-step derivations for the perfect annotator variance in our datasets.

\begin{table}[t]
    \centering
    \includegraphics[]{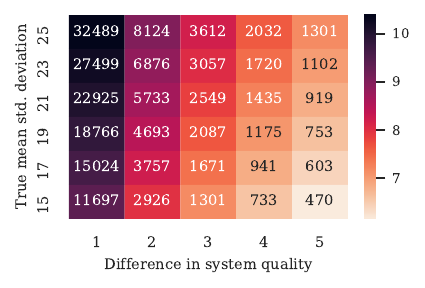}
    \caption{Power analysis for the number of judgments needed to give a pairwise prediction between two systems at $.9$ accuracy ($\alpha=0.05, \beta=0.95$) under ttest assumptions (normality, equal variance) in WMT. WMT ratings are on a 0-100 scale, and the perfect annotator variance in WMT19 was $19.27$. Darker cells indicate less feasible experiments, and the colors are set on a log scale. Results for SummEval are in Appendix \ref{appendix:power_analysis}.
    \label{table:wmt_power_analysis}}
\end{table}

\subsection{Results}

The following analyses refer to the comparison of metric estimators to unbiased estimators at varying number of judgments for WMT in Figure \ref{fig:human_comparisons}.

\textbf{Judgments from the perfect annotator have low variance, like those of professional linguists.} While we do not have data from professional linguists, we can qualitatively compare them to the perfect annotator. A growing body of MT literature focuses on professional linguists \cite{freitag-etal-2020-bleu, mathur-EtAl:2020:WMT}, and there are at least two known properties of their judgments: their judgments have better interannotator agreement (contain less noise), and they are more sensitive to linguistic phenomena. The perfect annotator has no noise, as they assign a constant score to each segment. However, the perfect annotator in WMT is better described as a noiseless crowdworker. With the biases of crowdworkers, the perfect annotator may not share the sensitivity property, and our use of crowdworkers may be biased w.r.t. professional linguists.

\textbf{In terms of average pairwise error, MT metrics have an equivalence to a high number of human judgments.} Since the error of the human estimator monotonically increases as the number of judgments decrease, each MT metric has a breakeven point. Metrics outperform human estimators using judgments below this threshold. {\sc BERTscore} is as accurate as using a human estimator with 600 judgments per system, or the perfect annotator estimator with 300 judgments, across the WMT dataset. We highlight the statistical advantage in variance many metrics share, and that this advantage offers a \emph{possibility} that metrics can outperform humans, determined by which human estimator the metric is compared against. This is a consequence of the general fact that humans are unbiased, high-variance estimators, and metrics are biased, low-variance estimators, as depicted in Figure \ref{figure:pairwise_illustration}. For metrics such as {\sc BERTscore} or {\sc chrF}, the bias is low as well, which gives it remarkably good error properties.


\section{The limits of human evaluation} \label{section:on_limits}
The perfect annotator provides optimistic figures for human annotation, providing the best performance for a fixed number of judgments, and requiring the least judgments for a fixed performance. In \S \ref{section:comparing_to_humans}, we saw that the perfect annotator is weak at low number of judgments, due to its non-zero variance. In this section we identify another consequence of the perfect annotator's variance, where estimating small differences in system quality is hard.

\subsection{Power analysis of the perfect annotator}

The performance of an unbiased estimator is dependent on their variance and the effect size it is trying to detect. This section performs a power analysis to determine how much annotator effort is needed to reliably detect the correct pairwise judgment between two systems 
\cite{card-etal-2020-little}. To make an optimistic estimate, we assume our annotator variance is close to that of a perfect annotator. We make two assumptions to apply a basic power analysis for the estimation of the difference of system quality between two systems: normality and equal variance across groups. For parameters $\alpha=0.05$ (false positive rate) and $\beta = 0.95$ (false negative rate), we can analytically compute the number of judgments needed to ensure our pairwise judgment is at least $\beta(1-\alpha)\approx90\%$ accurate. Table \ref{table:wmt_power_analysis} contains power analyses for different instantiations of annotator variance and effect size.

{\bf In WMT, detecting a difference of 1 point requires at least 10K perfect annotator judgments, for different instantiations of its variance.} To put this in perspective, the top 5 \texttt{zh-en} translation systems in WMT19 differed by less than 3 points \cite{barrault-etal-2019-findings}. Depending on how much is paid per judgment, this cost can quickly become infeasible. Here, the merit of such a task may be argued, as knowing a small difference exists between two systems may not always be productive. From a scientific perspective, many NLG techniques will yield small improvements, and not being able to detect small differences means we will not know whether these techniques are useful.

\subsection{Metrics more easily achieve significance}



Since metrics tend to have lower variance, metrics often achieve significance in estimating the difference of system qualities, when humans cannot. For instance, {\sc BERTscore} achieves significance in estimating quality differences over half of the pairwise examples where humans do not (see Appendix \S \ref{appendix:metric_human_significance}). {\bf In extreme cases, human evaluation is nearly as bad as flipping a coin, but the metric can still offer a consistent prediction between two systems.} When comparing systems similar in quality, practitioners must accept that the number of possible analyses are limited. In ablation studies where similar systems are often compared, metrics may be our only insight into system performance. With white-box metrics such as {\sc BLEU}, value can be derived from qualitative insight (e.g. systems with high {\sc BLEU} score have high $n$-gram overlap with the reference set). In addition, we may qualitatively analyze output statistics not intended to correlate with humans judgment at all \cite{neubig-etal-2019-compare}.

\section{Caveats to the analysis}

Our analysis assumes that the human judgments are unbiased. In WMT16-19, direct assessment \cite{graham-etal-2013-continuous} was used to elicit judgments from a combination of crowdworkers and researchers. Direct assessment (DA) uses an adequacy evaluation prompt (``Rate how much you agree that the output translation adequately expresses the meaning of the reference translation'') and asks contributors to rate on a 0-100 scale.

The unbiased ground truth is not a fixed goalpost. A number of factors are known to change the eventual ranking of translation systems with human scoring. Employing a different collection methodology, such as human translation edit rate (HTER) of instead of DA, can result in divergent system rankings \cite{graham-etal-2016-glitters}. In an earlier edition of WMT, DA judgments were collected with both a grammaticality prompt and an adequacy prompt, corresponding to different system rankings by the respective attribute \cite{Bojar2016TenYO}. Several studies have shown scoring differences between professional linguists and crowdworkers which are due in part to the fact that linguists are more sensitive to linguistic phenomena \cite{fabbri-etal-2019-multi, freitag-etal-2019-ape}.

The goals of an evaluation should be decided by the practitioner. We do not give suggestions on any particular goals, and practitioners should understand what their application is, and which evaluation is the best approximation \cite[refer to][]{DBLP:journals/jair/GattK18}. Unfortunately, since the existing data in this domain is limited, our analyses are limited as well. However, the statistical techniques apply to any empirical method. We hope that our analysis inspires others to think about statistical limits in this domain.


\section{Pushing the limits of evaluation}

To push the limits of what can be evaluated, we need to improve on fundamental aspects of human evaluation. On the human side, we may focus on creating larger effect sizes or reducing noise by adopting new annotation schemes \cite{laubli-etal-2018-machine, shapira-etal-2019-crowdsourcing} or employing professional linguists \cite{fabbri2020summeval, toral-EtAl:2018:WMT}. To make the human estimator more efficient, we may consider adaptive data collection techniques to stop data collection early when significance is achieved, in a statistically sound manner \cite{DBLP:conf/kdd/JohariKPW17}. 



Strategies combining human and metric evaluation are also shown to have potential. Variance reduction techniques can be applied to the human estimator by taking advantage of strong metrics \cite{chaganty-etal-2018-price}. Another bottleneck in human evaluation is in the random sampling of the test set. Metrics could form the basis of an importance sampling procedure to choose test sets that would best differentiate two systems, as a form of robust evaluation \cite{chaganty-etal-2017-importance}.

On the metric side, if we can reliably estimate metric bias, we can skip human evaluation altogether when the metric is known to be good. Probabilistic reinterpretations of current metrics could be a useful technique for confidence estimation \cite{keith-oconnor-2018-uncertainty}. Optimistically, metrics could have provable guarantees, ensuring the correctness of metric decisions \cite{jia-etal-2019-certified}.

\section{Best practices for metrics research}

We reinterpret problems in evaluating metrics with correlation (\S 2.2) as a set of guidelines for metrics research. To next year's organizers of the WMT metrics shared task and the broader metrics community we suggest the following: \textbf{(1)} Pairwise accuracy has desirable properties as an evaluation measure for metrics. Our bias-variance-noise decomposition shows that the observed pairwise accuracy is very close to the true pairwise accuracy from a noise-free, infinite test set setting (\S 4.4). We suggest the use of pairwise accuracy as it reflects metric performance well (which may be verified using this analysis). As a normalized form of pairwise accuracy, Kendall's $\tau$ is also a suitable measure. \textbf{(2)} Since pairwise accuracy is computed against noisy human predictions, on average, it should be impossible for metrics to achieve a perfect accuracy. We suggest providing an upper bound of metric performance (\S 4.2) to clarify how much improvement is possible for metrics on the dataset.


\section{Related work}

 The fact that a manual evaluation can be weak, and an automatic one can be better is gaining attention in the metrics community. \citet{mathur-EtAl:2020:WMT} studied a disagreement between crowdworkers and metrics, and a reevaluation favored the metrics over the human prediction. Recently, \citet{freitag2021experts} shows that metrics can achieve higher agreement with professional linguists than crowdworkers in judging translation systems. Their results fit into our formalization: if we assume professional linguists are unbiased, the bias and variance properties of metrics combined are superior to those of crowdworkers. Our analysis assumes that crowdworkers are unbiased, where they assume professional linguists are instead.

We wish to highlight several works which inspired the elements of ours: 
\citet{chaganty-etal-2018-price} and \citet{hashimoto-etal-2019-unifying} formalize metrics as statistical estimators and provide understanding of their statistical properties and limits. In the replication of ImageNet, \citet{DBLP:journals/corr/abs-2005-09619} found that dataset bias accounted for classifier performance differences between the original and the replicated dataset, and provide a decomposition for the sources of error. In automated essay scoring, scorers are often evaluated against noisy human judgment, and \citet{loukina-etal-2020-using} developed the PRMSE to calculate the MSE between scorer prediction and the true judgment, rather than noisy judgment. Finally, in bioinformatics, \citet{li2020performance} derive an upper bound of the $R^2$ coefficient due to experimental noise when regressing on experiment-derived results.

\section{Conclusion}

Through rigorous comparison between metrics, humans, and the perfect segment-level annotator, we identify the settings where metrics outperform humans due to a statistical advantage in variance. These results challenge the notion that metrics are always secondary to human evaluation. Instead, we encourage practitioners to understand when human evaluation is weak, and when metrics are necessary. Finally, we hope to provide tools for analysis and future directions for evaluation.

\section*{Acknowledgments}
Discussions with Nitika Mathur, Markus Freitag, and Thibault Sellam led to several insights. Nelson Liu and Tianyi Zhang provided feedback on our first draft, and anonymous reviewers provided feedback on the submitted draft. Nanyun Peng advised the first author, and on this work. Alex Fabbri provided a scored version of the SummEval dataset. We thank all who have made our work possible.

\bibliographystyle{acl_natbib}
\bibliography{anthology,acl2021}

\appendix
\newpage

\section{Equivalence between optimal prediction and true system differences} \label{appendix:optimal_predictions_proof}

There is a slight difference between the definition of the true difference in \eqref{eq:true_diff_label} which we can alternatively define as
\begin{equation} \label{eq:cond_true}
    \Delta^H = \text{sign}(\Expect[\widehat{\mu^H_{S}} - \widehat{\mu^H_{S'}}])
\end{equation}
and the definition of the optimal prediction $\Delta^{H*}$ in \eqref{eq:optimal_prediction}, which is positive when
\begin{equation} \label{eq:cond_optimal}
    P_{\mathcal{X}}(\widehat{\mu^H_{S}} - \widehat{\mu^H_{S'}} > 0) > \frac{1}{2}
\end{equation}
and the two are not immediately equivalent. However, if we assume that the central limit theorem applies (which can be reasonable as our sample means always have $n>100$) and $X = \widehat{\mu^H_{S}} - \widehat{\mu^H_{S'}}$ is normal, the CDF of $X$ is
\begin{equation}
    F(x) = \phi((x - \Expect[X]) / \Var(X))
\end{equation}
where $\phi$ is the CDF of the standard normal distribution. Since the standard normal is centered and symmetric, $\phi(x) > 1/2 \iff x > 0$. Together we have
\begin{equation}
    F(x) > \frac{1}{2} \iff \Expect[X] > x
\end{equation}
where for $x=0$ the left and right hand sides are equivalent to \eqref{eq:cond_optimal} and \eqref{eq:cond_true}, respectively.

\section{Convergence of metric predictions to the main prediction} \label{section:metric_convergence}

A key assumption in interpreting the results from the bias-variance-noise decomposition in \S \ref{section:decomposing_error} is that as system-level metrics have access to more outputs for evaluation, metric predictions converges onto the main prediction.

For many metrics, their system-level score is the mean of their segment-level scores (e.g. {\sc BLEURT}, {\sc BERTscore}, {\sc ROUGE} etc.). This is true for all summarization metrics. For these metrics, assuming the central limit theorem allows us to prove that metrics converge to the main prediction, similar to the proof in Appendix \ref{appendix:optimal_predictions_proof}. However, some MT metrics ({\sc BLEU}, {\sc TER}, and {\sc chrF}) are not simple averages of their segment-level scores, making them harder to analyze.

\begin{figure}[!h]
    \centering
    \includegraphics{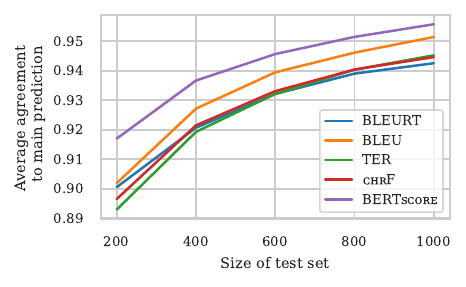}
    \caption{Average agreement of the main prediction to metric predictions computed from varying test set sizes in WMT. The main predictions were derived from all of our data. Each point was an estimated with 10K bootstrap trials. As the size of the test set increases, we see that the agreement monotonically increases. Note that only {\sc BLEURT} and {\sc BERTscore } are means of their segment-level scores. }
    \label{figure:metric_convergence_wmt}
\end{figure}

\begin{figure}[!h]
    \centering
    \includegraphics{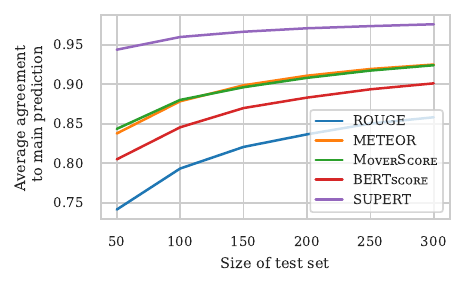}
    \caption{Average agreement of the main prediction to metric predictions evaluated on varying test set sizes in SummEval. The main predictions were derived from all of our data. Each point was an estimated with 10K bootstrap trials. As the size of the test set increases, we see that the agreement monotonically increases. Note that all metrics are means of their segment-level scores. }
    \label{figure:metric_convergence_summeval}
\end{figure}

For system-level metrics that are not simple averages, we analytically observe that their aggregation method is similar to a mean (e.g. {\sc BLEU} is a macro-average). We empirically verify that as the system-level metric evaluates on more test set outputs, their pairwise predictions converge to the main predictions. Refer to Figures \ref{figure:metric_convergence_wmt} and \ref{figure:metric_convergence_summeval}.

\newpage

\section{Efficiency ratios for the perfect annotator} \label{appendix:perfect_annotator_derivation}

With repeat human judgments for a given output example, we can estimate the variance of the  perfect annotator (or true segment-level score variance) in WMT and SummEval. For WMT, we use only valid judgments ($\texttt{SYSTEM}$ and $\texttt{REPEAT}$ judgments), and discard all attention check judgments ($\texttt{BAD\_REF}$ judgments). For SummEval, we use the dataset as is.

In WMT, we analyze all the to-English data grouped by year. We believe this grouping is appropriate because the to-English evaluation is batched together every year. Direct assessment, which WMT uses to collect human judgments \cite{graham-etal-2013-continuous}, is a score assigned by crowdworkers to an English translation while referring to an English reference, requiring only monolingual knowledge.

\begin{table}[!h]
    \small
    \centering
    \begin{tabular}{l|cccc}
         & 2016 & 2017 & 2018 & 2019 \\
         \midrule
        $\sqrt{\Var(H(x))}$ & 30.01 & 29.65 & 28.21 & 28.81 \\
        $\sqrt{\Expect[\Var(H(x) | x)]}$ & 17.53 & 22.96 & 19.57 & 21.42 \\
        $\sqrt{\Var(P(x))}$ & 24.36 & 18.76 & 20.33 & 19.27 \\
        $\Var(H(x)) / \Var(P(x))$ & 1.52 & 2.50 & 1.93 & 2.24 \\
    \end{tabular}
    \caption{Step-by-step derivation for the efficiency ratio $r$ (fourth row) of the perfect annotator estimator for WMT16-19 as defined in \S 4.1. Square roots are taken so that values are in terms of the original units (standard deviations, judgments range from 0-100). These were calculated on to-English data only. } \label{perfect_annotator_derivation_wmt}
\end{table}

\begin{table}[!h]
    \small
    \centering
    \begin{tabular}{l|cc}
         & Expert & Turker \\
         \midrule
$\sqrt{\Var(H(x))}$ & 0.717 & 0.745 \\
$\sqrt{\Expect[\Var(H(x) | x)]}$ & 0.293 & 0.475 \\
$\sqrt{\Var(P(x))}$ & 0.655 & 0.574 \\
$\Var(H(x)) / \Var(P(x))$ & 1.201 & 1.686 \\
    \end{tabular}
    \caption{Step-by-step derivation for the efficiency ratio $r$ (fourth row) of the perfect annotator estimator for SummEval as defined in \S 4.1. Square roots are taken so that values are in terms of the original units (standard deviations, judgments range from 1-5). Note that there is little agreement between experts and turkers at the system level.} \label{perfect_annotator_derivation_summeval}
\end{table}

\newpage
\section{SummEval analysis results} \label{appendix:power_analysis}

The main analyses in \S \ref{section:comparing_to_humans} and \S \ref{section:on_limits} are presented for SummEval here. When comparing expert humans to metrics, no metric comes close to expert performance at any number of expert judgments. For the power analysis, small differences are also hard to detect, similar to the findings in WMT. Note that while the perfect expert requires relatively less judgments compared to the perfect crowdworker in WMT, judgments from experts are likely to be much more expensive.

\begin{figure}[!h]
    \includegraphics{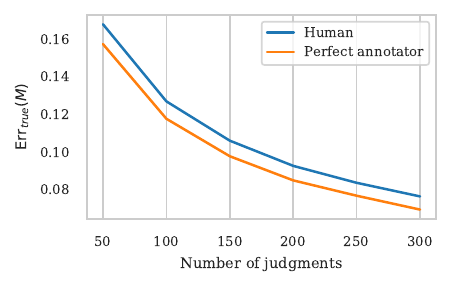}
\caption{Comparison of metrics to human and perfect annotator estimators with varying number of judgments in SummEval. Errors are adjusted to an idealized setting where true predictions are used for evaluation and metrics are computed on infinite test sets; here metric predictions become constant, so their errors are constant. No metric comes close to expert performance at any number of judgments ({\sc ROUGE}, the best performing summarization metric, has error $0.221$). }
\end{figure}

\begin{table}[!h]
    \centering
    \includegraphics[]{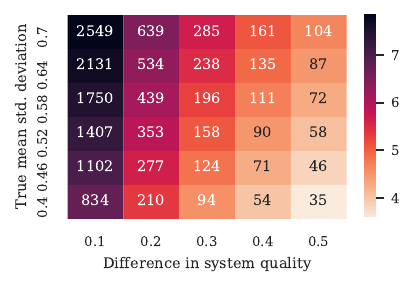}
    \caption{Power analysis for the number of judgments needed from the perfect expert to give a pairwise judgment between two systems at $.9$ accuracy ($\alpha=0.05, \beta=0.95$) under ttest assumptions (normality, equal variance) in SummEval. SummEval ratings are on a 1-5 scale, and the true segment quality variance was $0.655$. Darker cells indicate less feasible experiments, and the colors are set on a log scale.}
\end{table}

\newpage
\section{Metric and human significance breakdown} \label{appendix:metric_human_significance}

\begin{figure}[!h]
    \centering
    \includegraphics{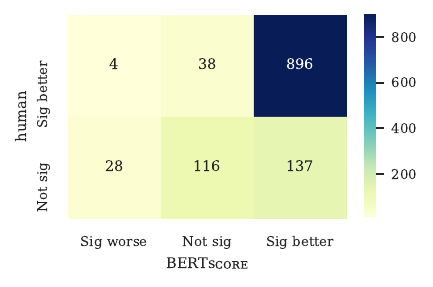}
    \caption{Co-occurrence of {\sc BERTscore} and human significance on pairs in WMT16-19. Pairs are ordered so that the human difference in system quality is always positive. Significance is tested with a one-sided bootstrap resampling test, in the direction of the difference for both humans and metrics with 1K trials at $\alpha=0.05$. {\sc BERTscore} achieves significance more than half of the time when humans cannot.}
    \label{figure:metric_human_significance_bert_score}
\end{figure}

\begin{figure}[!h]
    \centering
    \includegraphics{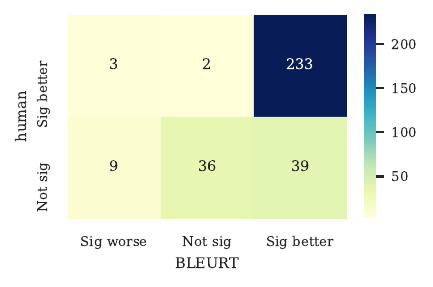}
    \caption{Co-occurrence of {\sc BLEURT} and human significance on pairs in WMT19. Pairs are ordered so that the human difference in system quality is always positive. Significance is tested with a one-sided bootstrap resampling test, in the direction of the difference for both humans and metrics with 1K trials at $\alpha=0.05$. {\sc BLEURT} achieves significance more than half of the time when humans cannot.}
    \label{figure:metric_human_significance_bleurt}
\end{figure}

For the pairwise examples in WMT, we provide the co-occurrence of significance for metric and human estimators. Refer to Figures \ref{figure:metric_human_significance_bert_score} and \ref{figure:metric_human_significance_bleurt} for analyses on {\sc BERTscore} and {\sc BLEURT}, respectively.

\end{document}